\documentclass{article}

\usepackage[utf8]{inputenc} % allow utf-8 input
\usepackage[T1]{fontenc}    % use 8-bit T1 fonts
\usepackage{hyperref}       % hyperlinks
\usepackage{url}            % simple URL typesetting
\usepackage{booktabs}       % professional-quality tables
\usepackage{amsfonts}       % blackboard math symbols
\usepackage{nicefrac}       % compact symbols for 1/2, etc.
\usepackage{microtype}      % microtypography
\usepackage{xcolor}         % colors

\usepackage{fullpage}
\usepackage{natbib}
\usepackage{amsmath}
\usepackage{amssymb}
\usepackage{amsthm}
\usepackage{bbm}
\usepackage{graphicx}
\usepackage{subcaption}
\usepackage{cleveref}
\usepackage{algorithm}
\usepackage{algorithmic}
\usepackage{notations}

\RequirePackage{array}
\newenvironment{authors}[1]%
  {\begingroup
   \newcommand\estyle{}%
   \newcommand\institute[1]%
     {\\\multicolumn{#1}{@{}c@{}}{\begin{tabular}[t]{@{}>{}c@{}}##1\end{tabular}}}%
   \newcommand\email[1]%
     {\gdef\estyle{\small\ttfamily}\\##1\gdef\estyle{}}
   \begin{tabular}[t]{@{}*{#1}{>{\estyle}c}@{}}
  }%
  {\end{tabular}%
   \endgroup
  }
\title{A Note on Optimizing Distributions \\ using Kernel Mean Embeddings}

\author{
\begin{authors}{3}% \\
  Boris Muzellec $^{* \star}$ & Francis Bach $^{* \star}$ &  Alessandro Rudi $^{* \star}$
  \email{boris.muzellec@inria.fr & francis.bach@inria.fr & alessandro.rudi@inria.fr}
\end{authors}
}

\date{\small{$^*$ INRIA Paris, 2 rue Simone Iff, 75012, Paris, France \\
			$^\star$ ENS - Département d’Informatique de l’École Normale Supérieure, \\
		    $^\star$ PSL Research University, 2 rue Simone Iff, 75012, Paris, France}}

\begin{document}

\maketitle

\begin{abstract}

Kernel mean embeddings are a popular tool that consists in representing probability measures by their infinite-dimensional mean embeddings in a reproducing kernel Hilbert space. When the kernel is characteristic, mean embeddings can be used to define a distance between probability measures, known as the {\em maximum mean discrepancy} (MMD). A well-known advantage of mean embeddings and MMD is their low computational cost and low sample complexity. However, kernel mean embeddings have had limited applications to problems that consist in optimizing distributions, due to the difficulty of characterizing which Hilbert space vectors correspond to a probability distribution. In this note, we propose to leverage the kernel sums-of-squares parameterization of positive functions of \cite{marteau2020non} to fit distributions in the MMD geometry. First, we show that when the kernel is characteristic, distributions with a kernel sum-of-squares density are dense. Then, we provide algorithms to optimize such distributions in the finite-sample setting, which we illustrate in a density fitting numerical experiment.

\end{abstract}

\section{Introduction}\label{sec:intro}

Mean embeddings~\citep{muandet2016kernel} are a way of representing probability distributions through the moments of a potentially infinite-dimensional feature vector, usually corresponding to the feature map $\phi(x)$ of a reproducing kernel Hilbert space (RKHS) $\hh$. When this RKHS is large enough (i.e., when the kernel is {\em characteristic}~\citep{sriperumbudur2011universality}) this embedding is injective, i.e., a distribution is uniquely characterized by its mean embedding. From there, one may define a distance between probability distributions as the distance between their embeddings in the Hilbert space, known as the maximum mean discrepancy (MMD)~\citep{gretton2012kernel}.

MMD benefits from a low computational cost and a favorable sample complexity. More precisely, given two distributions $\mu, \nu$ on $\RR^d$, one may get a estimate of the MMD distance based on $n$ samples from $\mu$ and $\nu$ with a precision $O(n\smrt)$, independently from the dimension $d$, at a $O(d n^2)$ computational cost. For those reasons, MMD has become a popular distance in the machine learning community that has had applications to testing \citep{gretton2012kernel,sejdinovic2013equivalence} and generative modeling \citep{li2015generative,binkowski2018demystifying}, among others.

However, numerous machine learning applications such as density fitting~\citep{parzen1962estimation,silverman1986density} or distributionally robust optimization (DRO)~\citep{rahimian2019distributionally} place the focus on optimizing distributions themselves. Despite its practical advantages, MMD has had limited applications to those tasks, due to the difficulty of characterizing which functions in a RKHS are the mean embeddings of a probability distribution. Indeed, while $\hh$ corresponds to the space spanned by $\{ \phi(x), x \in \RR^d\}$, the set of mean embeddings $\ME$ is only the {\em convex hull} of this set, whose extreme points are the feature vectors $\phi(x), x \in \RR^d.$ This can be related to the pre-image problem \citep{kwok2004pre}: given an element $v$ of $\hh$, the pre-image problem consists in finding a point $x$ whose feature vector $\phi(x)$ is equal or close to $v$. In comparison, we aim here at finding a {\em distribution} $\mu$ such that $\Esp_\mu[\phi(x)]$ is close to $v$.

As a workaround, \cite{staib2019mmddro} propose to relax such problems by optimizing over \emph{any} function in $\hh$, instead of restricting to those in $\ME $. As a result, the output of those methods are not guaranteed to correspond to probability distributions: they may correspond to measures that do not have unit mass, or that take negative values. Alternatively, in some settings it is possible to obtain a tractable exact problem by deriving the dual~\citep{zhu2021kernel}. 

\paragraph{Contributions.} In this short note, we leverage the kernel sum-of-squares representation for positive functions proposed by \citet{marteau2020non} to design a method to optimize over probability distributions in the MMD distance. 
In the finite-sample setting, this parameterization can be approximated with a finite-rank positive-semidefinite (PSD) matrix. Based on this fact, we provide algorithmic tools to solve optimization problems over distributions in the MMD geometry. Finally, we illustrate our methods on a density fitting example.

\section{Background and notation}\label{sec:background}

\paragraph{Notation.}
For $n \in \NN, [n]$ denotes the set $\{1, ..., n\}$. We use lower-case fonts for vectors (e.g., $v$), and bold upper-case fonts for matrices (e.g., $\bB$). We denote inner products $\dotp{\cdot}{\cdot}$ to which we add a subscript: $F$ for the Frobenius product between matrices, $\hh$ for the Hilbert inner product on $\hh$, $\HS$ for the Hilbert-Schmidt product between bounded operators on $\hh$.  For a set $\Xcal$, $\Prob(\Xcal)$ denotes the space of probability distributions on $\Xcal$, and when applicable $\mathcal{AC}(\Xcal)$ is the subset of absolutely continuous probability distributions (i.e., that admit a density w.r.t.\ the Lebesgue measure on $\Xcal$). $\Ccal_0(\Xcal)$ denotes the set of real-valued continuous functions on $\Xcal$ that vanish at infinity.

\paragraph{Kernels and reproducing kernel Hilbert spaces (RKHS).}
We refer to \citet{steinwart2008support} and \citet{paulsen2016introduction} for a more complete covering of the subject. Let $\Xcal$ be a set and $k : \Xcal \times \Xcal \mapsto \RR$. $k$ is a positive-definite kernel if and only if for any set of points $x_1, ..., x_n \in \Xcal$, the matrix of pairwise evaluations $K_{ij} = k(x_i, x_j), i, j \in [n]$ is positive semi-definite.
Given a kernel $k$, there exists a unique associated reproducing kernel Hilbert space (RKHS) $\hh$, that is, a Hilbert space of functions from $\Xcal$ to $\RR$ satisfying the two following properties:
\begin{itemize}
    \item For all $x\in \Xcal$, $k_x \defeq k(x, \cdot) \in \hh$;
    \item For all $f \in \hh$ and $x\in \Xcal$, $f(x) = \dotp{f}{k_x}_\hh$. In particular, for all $x, x' \in \Xcal$ it holds $\dotp{k_x}{k_{x'}}_\hh = k(x, x')$.
\end{itemize}
A feature map is a bounded map $\phi : \Xcal \mapsto \hh$ such that $\forall x, x' \in \Xcal, \dotp{\phi(x)}{\phi(x')}_\hh = k(x, x')$. A particular instance is $x \mapsto k_x$, which is referred to as the canonical feature map.  
Many practical applications of RKHS theory can be cast as minimization problems of the form
\begin{align}\label{eq:generic_rkhs_pb}
    \underset{f \in \Hcal}{\min}~~L(f(x_1), ..., f(x_n)) + \Omega(\|f\|_\hh),
\end{align}
where $\Omega : \RR_+ \mapsto \RR$ is a strictly increasing function. In that case, the representer theorem \citep[see, e.g.,][and references therein]{steinwart2008support,paulsen2016introduction} shows that solutions of $\eqref{eq:generic_rkhs_pb}$ admit the following finite representation: $f = \sum_{i=1}^n \alpha_i k_{x_i}$ for some $\alpha \in \RR^n$. 

\paragraph{Kernel mean embeddings and maximum mean discrepancy (MMD).}

Given a probability measure~$\mu$, we define its kernel mean embedding as the element $w_\mu$ of $\hh$ that satisfies $\forall f \in \hh, \, \Esp_{X\sim \mu}[f(x)] = \dotp{f}{w_\mu}_\hh$, which may be expressed as $w_\mu = \int_\Xcal \phi(x) \dd\mu(x)$.
When the kernel is \emph{characteristic} \citep{sriperumbudur2011universality} -- such as the Gaussian kernel -- the embedding $\mu \mapsto w_\mu$ is injective. Note that mean embeddings are a strict subset of $\hh$: as mentioned in \Cref{sec:intro}, an element of $\hh$ is the kernel mean embedding of a distribution if and only if it lies in $\ME \defeq \Conv(\{\phi(x) : x \in \hh \})$, whereas $\hh = \Span(\{\phi(x) : x \in \hh \})$.
Using kernel mean embeddings, we may define a distance between probability measures, called the \emph{maximum mean discrepancy} \citep{gretton2012kernel}: 
$$\MMD(\mu, \nu) \defeq \|w_\mu - w_\nu\|_\hh.$$
When $\nu$ admits a density $p$, we write by abuse $\MMD(\mu, p)$ for $\MMD(\mu, \nu)$.

\subsection{Relaxing mean embedding constraints: a counter-example}\label{subsec:counter-example}
\begin{figure}[ht]
    \centering
    \includegraphics[width = .48\textwidth]{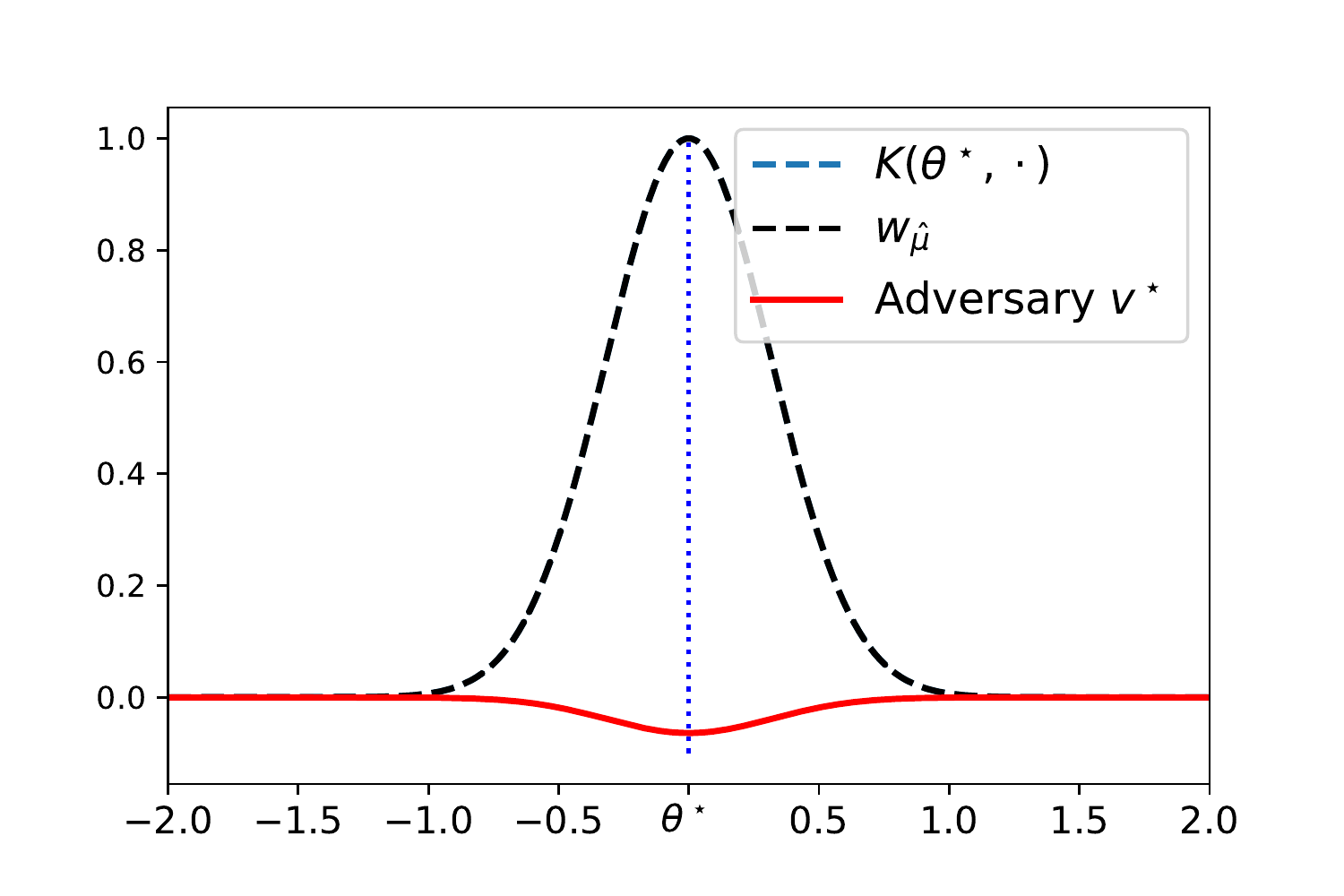}
    \includegraphics[width = .48\textwidth]{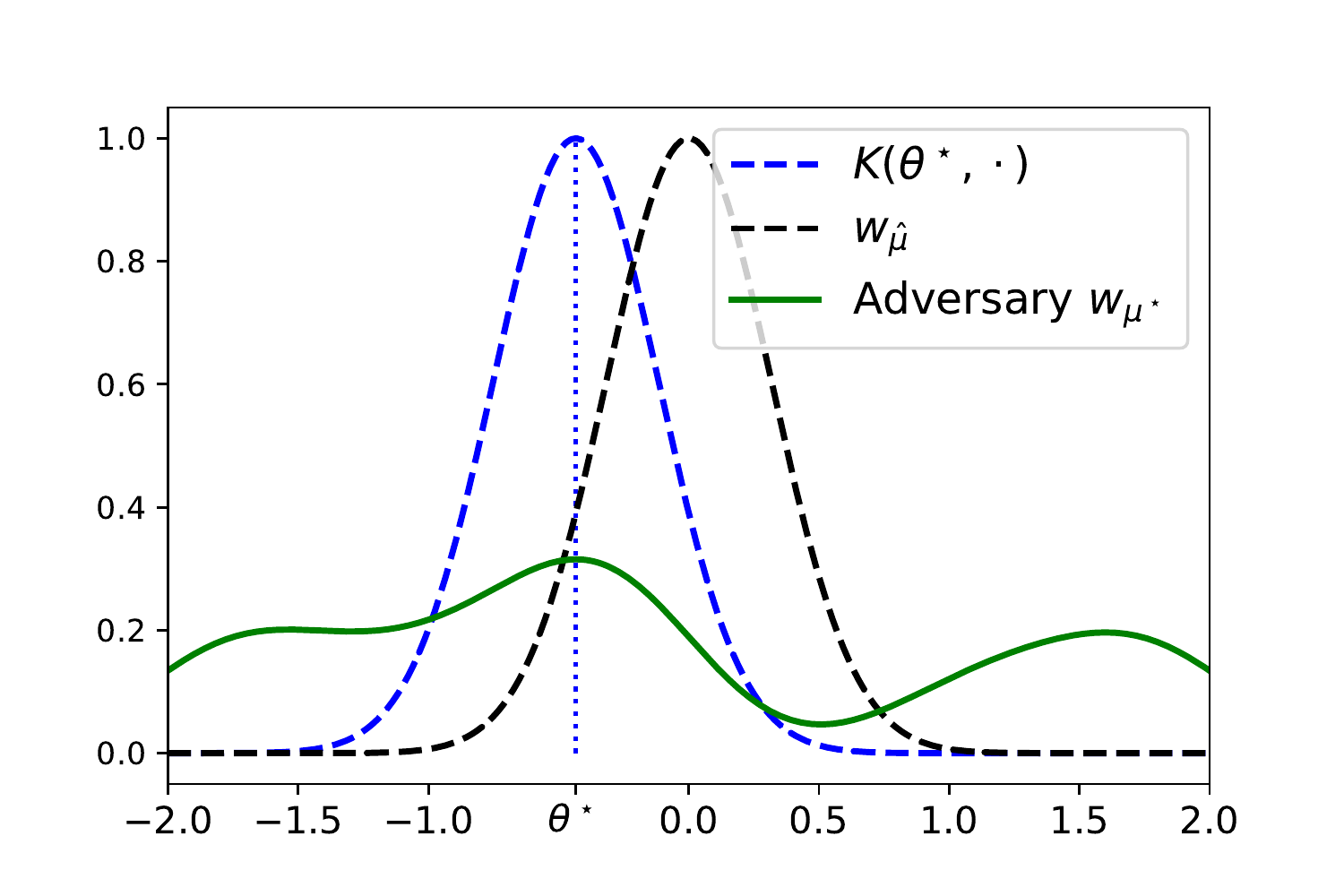}
    
    \vspace*{-.25cm}
    
    \caption{RKHS norm relaxation {\em(left)} vs. MMD solution {\em(right)}. Note that the relaxed adversary (left) is not the mean embedding of a probability distribution (it is the embedding of a Dirac in $0$ with a negative mass). The solution of the unrelaxed problem is evaluated by discretizing the support interval [-2, 2], for which we obtain $\theta^\star \approx -0.4$. A equivalent symmetrical saddle point with $\theta^\star \approx 0.4$ also exists.}
    \label{fig:example_mode_finding}
\end{figure}

We conclude this section with a simple example to illustrate how replacing MMD constraints over distributions with RKHS norm constraints over vectors can be detrimental. Let $\hat{\mu}$ denote a distribution of observed data points, and $K$ a positive-definite translation-invariant kernel (e.g., the Gaussian kernel $K(x, y) = e^{-\frac{\|x-y\|^2}{\sigma^2}}$) with corresponding RKHS $\hh$. Assume the convolution $(K \ast \hat{\mu})$ of $\hat{\mu}$ by $K$ denotes a ``return function'',
that a player wants to maximize by picking the highest mode:
\begin{align}\label{eq:mode_finding}
    \max_{\theta \in \Rd}~(K \ast \hat{\mu})(\theta).
\end{align}
The player knows that an adversary might have perturbed the data $\hat{\mu}$ that they have observed, compared to a true underlying distribution $\mu$ from which the player will get their returns. To hedge themselves against the adversary, the player decides to optimize $\theta$ over the worst distribution in a MMD ball around the observed distribution $\hat{\mu}$:
\begin{align}\label{eq:adv_mode_finding}
    \max_{\theta \in \Rd} \min_{\mu \in \Prob(\Rd)} (K \ast \mu)(\theta) ~~~\st~~~\MMD(\mu, \hat{\mu}) \leq \varepsilon.
\end{align}
As mentioned in the introduction, problems of this form are generally intractable. As an example, \citet{staib2019mmddro}  are confronted to this issue in applications to distributionally robust optimization, and propose to circumvent the difficulty by relaxing the problem: instead of restricting to distributions in the MMD distance, they optimize over $\hh$ in the RKHS norm.
Observing that the objective of \cref{eq:adv_mode_finding} can be rewritten as the  inner product $(K \ast \mu)(\theta) = \dotp{w_\mu}{\phi(\theta)}_\hh$, this yields the following relaxed problem:
\begin{align}\label{eq:relaxed_adv_mode_finding}
    \max_{\theta \in \Rd} \min_{v \in \hh}~~ \dotp{v}{\phi(\theta)}_\hh ~~~\st~~~\|v-w_{\hat{\mu}}\|_\hh \leq \varepsilon.
\end{align}
\Cref{eq:relaxed_adv_mode_finding} is a saddle point problem, in which the inner minimum a is convex problem. Writing the optimality conditions for the min, we get that the optimal adversary is $v^\star(\theta) = w_{\hat{\mu}} - \eps\frac{\phi(\theta)}{\|\phi(\theta)\|}$. Plugging this back in \cref{eq:relaxed_adv_mode_finding}, the problem becomes 
\begin{align}\label{eq:relaxed_adv_mode_finding_2}
\max_{\theta\in \RR^d} ~(K\ast\hat{\mu})(\theta) - \eps \sqrt{K(\theta, \theta)}.
\end{align}
Let us observe two things: first, in general $v^\star(\theta)$ is not the mean embedding of a probability distribution, and may even take negative values. Second, given that $K$ is a translation-invariant kernel, the solution $\theta^\star$ of \cref{eq:relaxed_adv_mode_finding_2} is the same as the non-robust version \eqref{eq:mode_finding} and does not depend on $\eps$: only the value of the objective does. Hence, the relaxed version \eqref{eq:relaxed_adv_mode_finding} fails at guaranteeing adversarial robustness. On the other hand, the original adversarial problem over distributions in \cref{eq:adv_mode_finding} does not admit a simple analytical expression, but does guarantee robustness against adversarial perturbations. This is illustrated in \Cref{fig:example_mode_finding} in a 1D case where $\hat{\mu}$ is a Dirac centered in 0 and $K$ is the Gaussian kernel. The discrepancy between the MMD geometry over distributions and the RKHS norm over general functions is already visible in this toy example. This suggests that the RKHS norm relaxation is ill-suited to more complex, higher-dimensional tasks and settings, and calls for better ways of dealing with optimization problems on distributions with MMD.

\section{Optimizing over distributions using kernel sums-of-squares}
As mentioned in \Cref{sec:intro}, problems of the form  
\begin{align}\label{eq:generic_mmd_pb}
    \inf_{\mu \in \Prob(\Xcal)} F(v) ~~\st~ v = \int \phi(x) \dd\mu(x)
\end{align}
are notoriously difficult to tackle. In this note, we introduce a parametric model for smooth distributions that is compatible with MMD, that can be plugged in \cref{eq:generic_mmd_pb}.
We focus on measures that admit a smooth density $p$ w.r.t.\ to a reference measure $\rho \in \Mcal(\Xcal)$ (e.g., for $\Xcal \subset \RR^d$ we may consider the Lebesgue measure), and propose to represent such densities using the kernel sum-of-squares (SoS) representation of non-negative functions of \cite{marteau2020non}:
\begin{equation}\label{eq:A_density}
    p_A(x) = \dotp{\phi(x)}{A \phi(x)}_\hh,~~~ x \in \Xcal, 
\end{equation}
with $A \in \pdm{\Hcal}$.  The following lemma (which is a particular case of Proposition 4 of \cite{marteau2020non}) characterizes the operators $\pdm{\Hcal}$ that lead to a valid density $p_A$ (i.e., with total mass equal to $1$). Its proof is deferred to the appendix.
\begin{lemma}\label{lemma:density}
    Let $A \in \pdm{\Hcal}$ and $\rho \in \Mcal(\Xcal)$. Define $\Sigma_\rho \defeq \int_\Xcal \phi(x)\otimes \phi(x) \dd\rho(x)$. The function $p_A$ defined in \cref{eq:A_density} is a density w.r.t.~$\rho$ if and only if $\dotp{A}{\Sigma_\rho}_\HS = 1$. 
\end{lemma}
From there, we propose to handle problems of the form \eqref{eq:generic_mmd_pb} using the parameterization in \cref{eq:A_density}, i.e., to solve 
\begin{align}
    \inf_{A \in \pdm{\hh}} F(w_{p_A}) ~~\st~~ \tr(A\Sigma_\rho) = 1.
\end{align}
We will denote $\Fcal_\rho \defeq = \{p_A: A\in \pdm{\hh}, \tr A\Sigma_\rho = 1\}$. This representation has the double advantage of being parametric -- and thus amenable to learning, as we will show in the remainder of this work -- and universal. Indeed, as proved by \cite{marteau2020non}, any continuous positive function can be approximated arbitrarily well in maximum norm over compact subsets by a function of the form of \cref{eq:A_density} provided $\hh$ is large enough: this is referred to as {\em universality} \citep{micchelli2006universal}. 
\begin{proposition}[\citet{marteau2020non}]\label{thm:marteau_universality}
    Let $\hh$ be a RKHS with a universal feature map $\phi : \Xcal \mapsto \RR^d$. Then $\{p_A: A\in \pdm{\hh}, \tr A < \infty\}$ is a universal approximator of continuous non-negative functions on $\Xcal$.
\end{proposition}
In particular, this representation allows to approximate continuous density functions on $\Xcal$ arbitrarily well. However, depending on the choice of topology, \Cref{thm:marteau_universality} alone does not guarantee that \emph{any} distribution can be approximated by a distribution with a density in $\Fcal_\rho$. For instance, in the total variation distance such densities may approach distributions with continuous density functions, but may not approach Dirac distributions. We show in the following theorem that distributions with densities in $\Fcal_\rho$ are dense in the set of probability distributions for the weak topology. When the kernel is continuous (like most usual kernels) and when $\hh \subset \Ccal_0$, this implies in turn that $\Fcal_\rho$ is dense for the MMD distance \citep[Lemma 2.1]{simon2020metrizing}.
\begin{theorem}\label{prop:weak_density}
    Let $\Xcal$ be a compact subset of $\RR^d$, $\hh$ be a RKHS with a universal feature map $\phi : \Xcal \mapsto \RR^d$, and assume $\rho$ is absolutely continuous. Then distributions with densities w.r.t.\ $\rho$ in $\Fcal_\rho$ are dense in $\Prob(\Xcal)$ for the topology of the weak convergence.
\end{theorem}

\paragraph{Examples.}
Kernels and RKHS satisfying the hypothesis of \Cref{thm:marteau_universality} and \Cref{prop:weak_density} include (but do not limit to):
\begin{itemize}
    \item the Gaussian kernel $k(x, x') = e^{-\frac{\|x-x'\|^2}{\sigma^2}}$;
    \item the Laplace kernel $k(x, x') = e^{-\frac{\|x-x'\|}{\sigma}}$;
    \item more generally, the Sobolev kernels $k_s(x, x') \propto \|x - x'\|^{s - d/2}\Kcal_{s-d/2}(\|x-x'\|)$ with $s > d/2$ and where $\Kcal_{s-d/2}$ is the Bessel function of the second kind, whose corresponding RKHS are the Sobolev spaces of smoothness $s$ \citep{adams2003sobolev}.
\end{itemize}

Finally, Hilbert space distances of mean embeddings with densities of the form \cref{eq:A_density} (and MMD in particular) can be expressed as functions of $\phi$-tensors of order $4$:

\begin{lemma}\label{lemma:rkhs_norm}
Let $A\in \pdm{\hh}, \tr A < \infty$ and $v \in \hh$. It holds 
\begin{align}
    \|v - w_{p_A}\|^2_\hh = \|v\|_\hh^2 + \dotp{A}{\Tcal (A)}_\HS - 2 \dotp{A}{\Vcal}_\HS,
\end{align}
\begin{align*}
    &\text{with}~~ \Vcal   \defeq \int v(x) \phi(x)\otimes \phi(x) \dd \rho(x)\\
 &\text{and}~~\Tcal (A) \defeq \iint \phi(x)\otimes \phi(x) \dotp{\phi(y)}{A\phi(y)}_\HS\, k(x,y) \dd \rho(x) \dd\rho(y) \ \in\  \pdm{\hh}.
\end{align*}
\end{lemma}

\subsection{Low-rank representations}

As recalled in \Cref{sec:background}, a key benefit of working in an RKHS is the existence of the representer theorem, which allows to learn functions from samples in a finite-dimensional representation. \citet{marteau2020non} prove an extension of the representer theorem for kernel SoS of the form \eqref{eq:A_density}. More precisely, the solution of a problem of the form
\begin{align}
\underset{A \in \pdm{\hh}}{\min} L(p_A(x_1), ..., p_A(x_n)) + \lambda \tr A
\end{align}
admits a representation of the form $A = \sum_{ij}^n B_{ij}\phi(x_i)\otimes \phi(x_j)$ with $\bB \in \pdm{\RR^n}$. However, unless $\Sigma_\rho$ admits a finite-rank parameterization, this result does not hold under the additional constraint $\dotp{A}{\Sigma_\rho}_\HS = 1$ that is required to ensure that $p_A$ is a density (\Cref{lemma:density}).

As a workaround, we make an approximation and consider problems between vectors that are projected on a finite-dimensional subspace $\hh_m = \Span\{\phi(\tx_1), ..., \phi(\tx_m)\}$ where $\tx_1, ..., \tx_m$ are a set of supporting points, and consider the case where $A \in \pdm{\hh_m}$. In that case, we can write $A = \sum_{i=1}^m B_{ij}\phi(\tx_i)\otimes \phi(\tx_j)$ with $\bB \in \pdm{\RR^m}$, and we define $p_\bB(x) = \sum_{ij=1}^m B_{ij}k(x, \tx_i) k(x, \tx_j)$. From \Cref{lemma:density}, $p_\bB$ is a valid density if and only if $\tr(\bB\bW) = 1$ with $W_{ij} \defeq \int_\Xcal k(x, \tilde{x}_i) k(x, \tilde{x}_j)\dd\rho(x), i, j \in [m]$.
Let $P_m$ denote the orthogonal projection onto $\hh_m$, i.e. $P_m(\phi(x)) = \sum_{i=1}^m c_i \phi(\tx_i)$ with $c = \tbK\inv \tk_x$ where $\tilde{K}_{ij} = k(\tx_i, \tx_j), i, j \in [m]$ and $(\tk_x)_i = k(x, \tx_i), i \in [m]$. In particular, for a function that admits a finite representation $v=\sum_{i=1}^n a_i\phi(x_i)$, we have $P_m(v) = \sum_{i=1}^m b_i\phi(\tx_i)$ with $b = \tilde{\bK}\inv \bK(\tilde{X}, X)a$ and $\bK(\tilde{X}, X)_{ij} = k(\tx_i, x_j), i\in [m], j\in [n]$.
The following lemma gives a closed-form expression of the RKHS distance between a vector in $\hh$ that admits a finite representation, and the mean embedding $w_{p_\bB}$. Its proof is deferred to \Cref{sec:proofs}.
\begin{lemma}\label{lemma:mmd_closed_form}
Let $\bB \in \pdm{\RR^m}$ and $v=\sum_{i=1}^n a_i\phi(x_i)$, and $W_{ij} \defeq \int_\Xcal k(x, \tilde{x}_i) k(x, \tilde{x}_j)\dd\rho(x), i, j \in [m]$. For $p, q\in [m]$, denote $\bu_{pq} = \int k(x, \tx_p) k(x, \tx_q) \tk_x \dd\rho(x) \in \RR^m$. Then, $p_\bB$ is a density function if and only if $\tr(\bB\bW) = 1$,  and it holds
\begin{align}
     \|P_m(v - w_{p_\bB})\|^2_\Hcal = \Ucal(\bB)^T\tilde{\bK}\inv\Ucal(\bB) - 2\dotp{\bB}{\bV}_F + \|P_m(v)\|_\hh^2,
\end{align}
with $ \forall p, q \in [m], V_{pq} =  \bu_{pq}^T\tilde{\bK}\inv \bK(\tilde{X}, X) a$, and $\Ucal$ is the map from $\RR^{m\times m}$ to $\RR^m$ such that $\Ucal(\bB) = \sum_{1 \leq p,q\leq m} B_{pq}\bu_{pq}$.
\end{lemma}
In particular, whenever the vectors $\bu_{pq}, p, q \in [m]$ are available in closed form (see examples in \Cref{sec:closed_forms}), $\|P_m(v - w_{p_\bB})\|^2_\Hcal$ can be computed in $O(m^3 + nm^2)$ time.

\subsection{Algorithms}\label{subsec:algos}

We now provide algorithmic tools to optimize distributions with densities in $\Fcal_\rho$. We illustrate those techniques on the sample problem of fitting a distribution with a kernel SoS density to observed data, with an optional trace regularization term: 
\begin{equation}\label{eq:low_rank_pb}
    \inf_{A \in \pdm{\hh_m}} \|P_m(v - w_{p_A})\|^2_\Hcal + \lambda \tr A ~~\st~~ \tr (A\Sigma_\rho) = 1.
\end{equation}
Writing $A = \sum_{ij} B_{ij}\phi(\tx_i) \otimes \phi(\tx_j)$, we have $\tr A = \sum_{ij} B_{ij} k(\tx_i, \tx_j)$. Hence, from \Cref{lemma:mmd_closed_form} and ignoring terms that do not depend on $\bB$, \cref{eq:low_rank_pb} can be reformulated as 
\begin{align}\label{eq:low_rank_pb_trace}
    \inf_{ \bB \succeq 0} \Ucal(\bB)^T\tilde{\bK}\inv\Ucal(\bB) - \dotp{\bB}{2\tilde{\bV} -\lambda \tilde{\bK}}_F ~~\st~~ \tr(\bB\bW) = 1,
\end{align}
with $W_{ij} \defeq \int_\Xcal k(x, \tilde{x}_i) k(x, \tilde{x}_j)\dd\rho(x), i, j \in [m]$.
\Cref{eq:low_rank_pb_trace} is a smooth convex optimization problem, and can therefore be solved using (accelerated) gradient descent. However, the projection on the set $\Acal \defeq \{\bM \in \pdm{\RR^m} : \tr(\bM\bW) = 1\}$ is computationally expensive (it is an instance of the covariance adjustment problem~\citep{malick2004dual,boyd2005least}). 
We circumvent this issue by changing the parameterization to $\bC = \bR \bB \bR^T$, where $\bR$ satisfies $\bW = \bR^T\bR$ (e.g., the Cholesky factor of $\bW$). With this parameterization, we have 
$$\bB\in \Acal \iff \bC \in \Bcal  \defeq \{\bM \in \pdm{\RR^m} : \tr(\bM) = 1\},$$
with the advantage that the projection on $\Bcal$ is much easier to compute: see \Cref{alg:proj_B}. This projection relies on computing the eigenvalue decomposition of $\bC$ (in $O(m^3)$ time) followed by the projection of its eigenvalues on the simplex in $O(m)$ time~\citep{maculan1989linear}. 
With this parameterization, \cref{eq:low_rank_pb_trace} becomes
\begin{align}\label{eq:C_param_pb}
    \inf_{ \bC \succeq 0} \Ucal(\bR\inv\bC\bR^{-T})^T\tbK\inv\Ucal(\bR\inv\bC\bR^{-T}) +  \dotp{\bC}{\bR^{-T}(\lambda\tbK - 2\bV)\bR\inv}_F ~~\st~~ \tr(\bC) = 1.
\end{align}
 We may optimize this objective using projected gradient descent, or FISTA~\citep{beck2009fast}. The initial formation of $\bV$, $\bR\inv$ and $\bu_{pq}, p,q \in [m]$ has a $O(mn + m^3)$ computational cost, and $O(mn + m^3)$ memory footprint. From then, each (accelerated) projected gradient iteration has complexity $O(m^3)$ and $O(m^3)$ memory footprint (since storing $\bK(\tilde{X}, X)$ is not necessary once $\bV$ is formed). Hence, using FISTA \cref{eq:low_rank_pb_trace} can be minimized to precision $\varepsilon$ with a total $O\left(mn + \frac{m^3}{\varepsilon^2}\right)$ computational cost~\citep[][Theorem 4.4]{beck2009fast}.
\begin{algorithm}[ht]
\caption{Projection on $\Bcal$}\label{alg:proj_B}
\begin{algorithmic} 
\REQUIRE $\bX \in \sym{\RR^n}$
\STATE Compute the EVD of $\bX$: $\bX = \bU \diag(\lambda_1, ..., \lambda_n) \bU^T$.
\STATE Project on the simplex: $(\lambda_1, ..., \lambda_n)_+ \leftarrow \mathrm{proj}_{\Delta_n} (\lambda_1, ..., \lambda_n)$, \qquad $\Delta_n \defeq\{x \in \RR_+^n : \sum_{i=1}^n x_i = 1\}$
\ENSURE $\bX_+ =  \bU \diag(\lambda_1, ..., \lambda_n)_+ \bU^T$
\end{algorithmic}
\end{algorithm}

\begin{remark}
Depending on the choice of kernel parameters, \cref{eq:low_rank_pb_trace} and \cref{eq:C_param_pb} may have poor conditioning. In particular, forming the Cholesky decomposition $\bW = \bR^T\bR$ and inverting $\bR$ may suffer from numerical stability issues. While a classical way of dealing with such issues is to use pre-conditioning~\citep{rudi2017falkon} on $\bB$, this approach is not compatible with the projection strategy described in \Cref{alg:proj_B}. As an alternative, when the conditioning is problematic we propose to add a small diagonal term to $\bW$ and to apply the constraint $\tr(\bB(\bW + \lambda \eye)) = 1$, and then renormalize $\bB$ to satisfy \Cref{lemma:density}: $\bB\leftarrow\frac{\bB}{1 - \lambda \tr\bB}$.
\end{remark}

\section{Applications and numerical experiments}
 Motivated by \Cref{prop:weak_density}, we propose to handle problems of the form \eqref{eq:generic_mmd_pb} using our parameterization, i.e., to solve 
\begin{align}\label{eq:sos_param_pb}
    \inf_{A \in \pdm{\hh}} F(w_{p_A}) ~~\st~~ \tr(A\Sigma_\rho) = 1.
\end{align}

\paragraph{Example 1: density fitting.}
Given $n$ points $x_1, ..., x_n$ sampled from an unknown distribution $\mu$, density fitting~\citep{parzen1962estimation} aims at estimating a parametric model $\mu_\theta$ for $\mu$ based on its samples. Motivated by \Cref{prop:weak_density}, we propose to fit a density estimator $p_\bB \in \Fcal_\rho$ using MMD, with an optional regularization term:
\begin{equation}\label{eq:densit_fitting}
\min_{\bB \succeq 0}~~ \MMD(\frac{1}{n}\sum_{i=1}^n \delta_{x_i}, p_\bB) + \lambda \tr \bB\bK ~~\text{such that}~~\tr(\bB\bW) = 1.
\end{equation}
\paragraph{Example 2: distributionally robust optimization.}
Kernel distributionally robust optimization (DRO) \citep{staib2019mmddro,zhu2021kernel} consists in minimizing a loss function $\ell_f$ over $f\in \hh$ under bounded adversarial perturbations of the input data $\hat{\mu} = \sum_{i=1}^n\delta_{x_i}$:
\begin{align}\label{eq:kernel_DRO}
    \begin{split}
    \min_{f \in \hh} \max_{\mu \in \Prob(\Xcal)} & \Esp_{x\sim \mu}[\ell_f(x)] ~~\st~~ \MMD(\mu, \hat{\mu}) \leq \varepsilon.
    \end{split}
\end{align}
Due to the difficulty of optimizing under the constraint $\MMD(\mu, \hat{\mu}) \leq \varepsilon$, several relaxations of this problem have been proposed, leading to problems that do not respect the $\MMD$ geometry, as illustrated in \Cref{subsec:counter-example}.
We propose instead to perform DRO with the parameterization \eqref{eq:sos_param_pb}. Problem \eqref{eq:kernel_DRO} then becomes
\begin{align}\label{eq:sos_DRO_penalized}
    \begin{split}
    \min_{f \in \hh} \max_{\bB \in \pdm{\RR^n}} & \sum_{i,j=1}^n B_{ij} \int \ell_f(x) k(x, x_i) k(x, x_j) \dd \rho(x)  \\
    \textrm{subject to}~~~ &\tr(\bB\bW) = 1 ~~\text{and}~~ \MMD(\hat{\mu}, p_\bB) \leq \varepsilon.
    \end{split}
\end{align}
Provided $\ell_f$ is convex in $f$, \eqref{eq:kernel_DRO} is a convex-concave min-max problem. Hence, whenever the integral $\int \ell_f(x) k(x, x_i) k(x, x_j) \dd \rho(x)$ can be computed in closed form (e.g., for the square loss with a Gaussian kernel), \cref{eq:kernel_DRO} can be solved using standard min-max optimization techniques. 
\paragraph{Numerical experiments.}
\begin{figure}[ht]
    \centering
\begin{subfigure}[b]{1.\textwidth}
    \includegraphics[width = .32\textwidth]{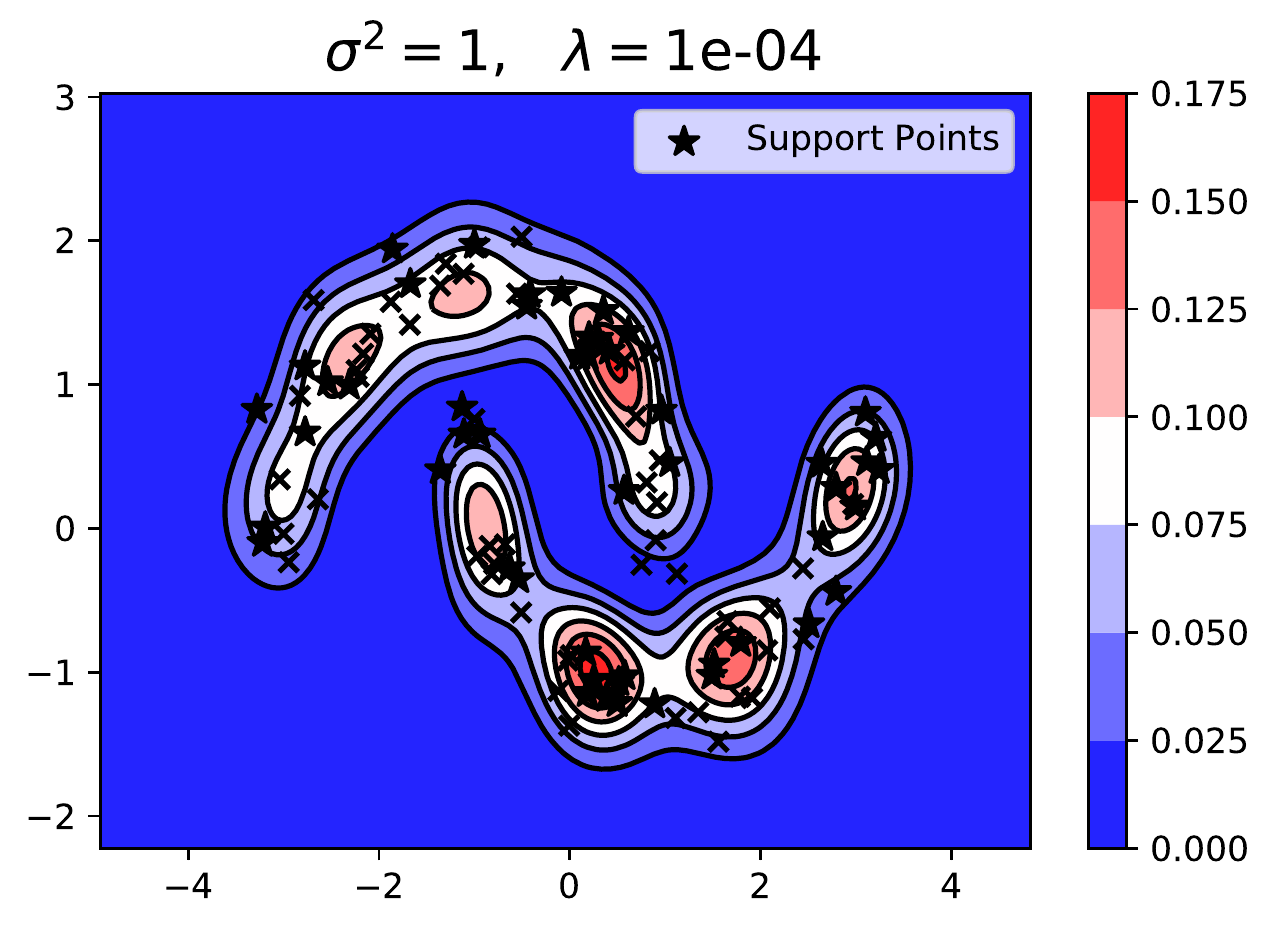}
    \includegraphics[width = .32\textwidth]{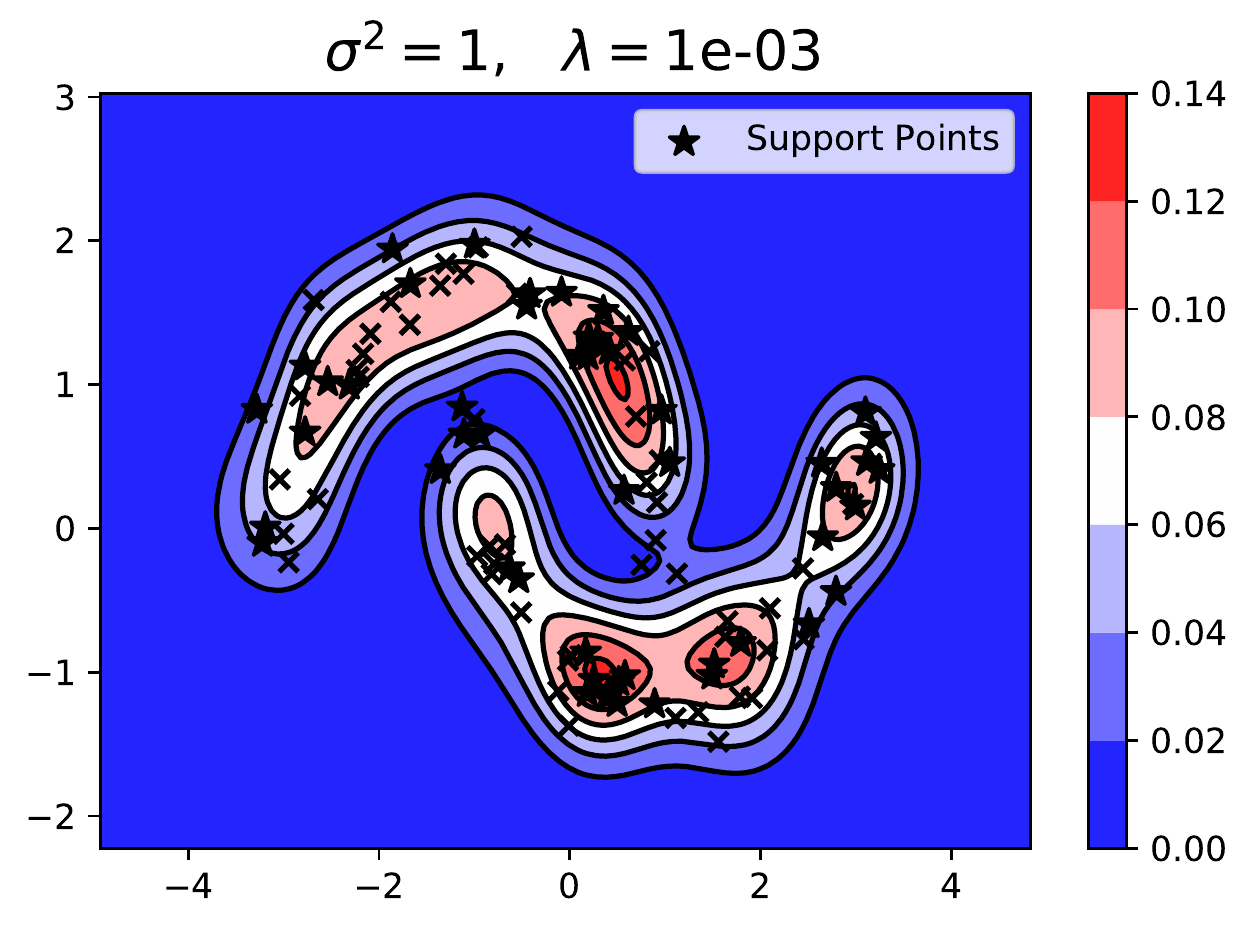}
    \includegraphics[width =  .32\textwidth]{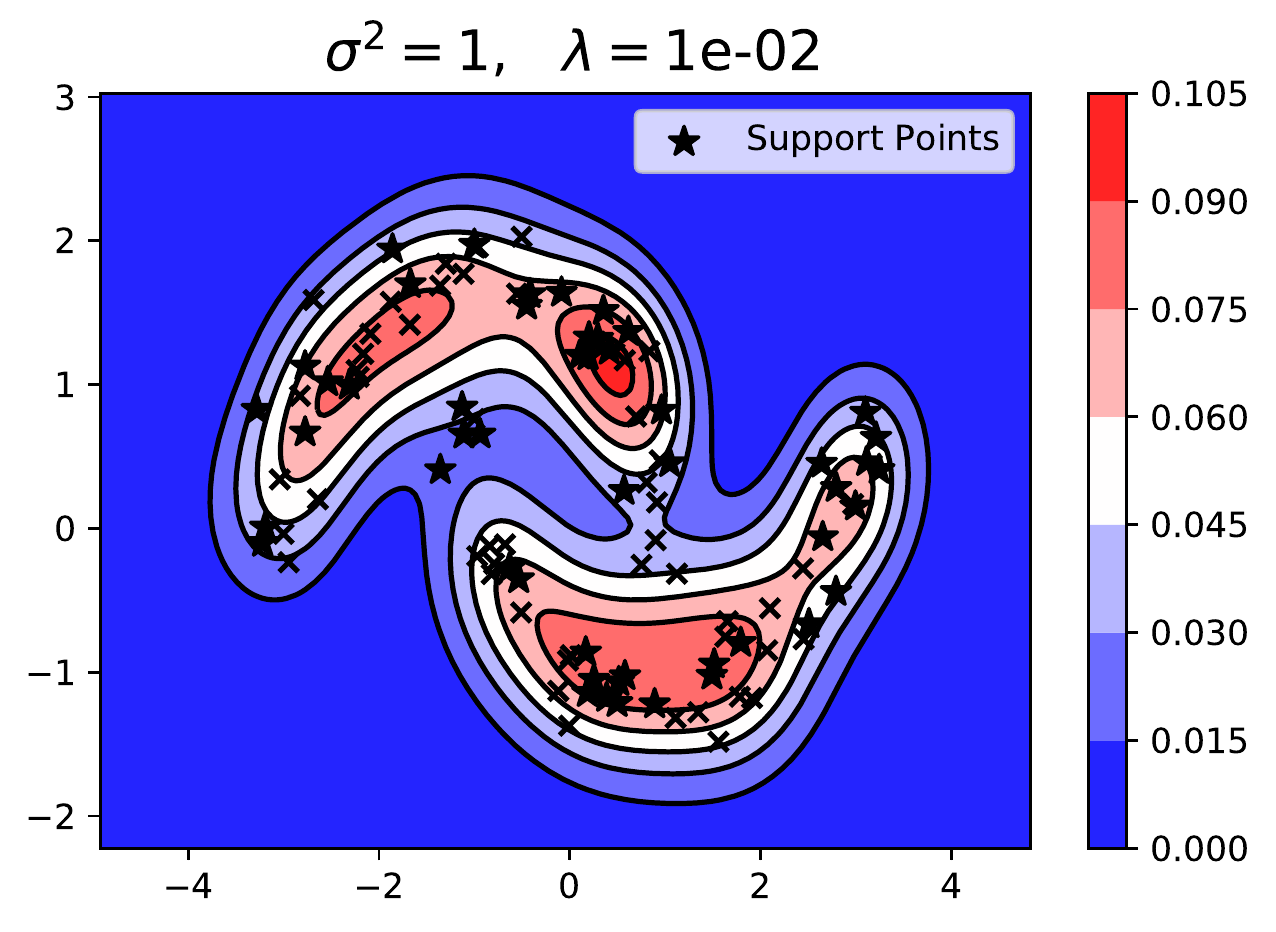}
    \caption{Effect of regularization.}
    \end{subfigure}
    \label{fig:lbda_effect}
\begin{subfigure}[b]{1.\textwidth}
    \centering
    \includegraphics[width = .32\textwidth]{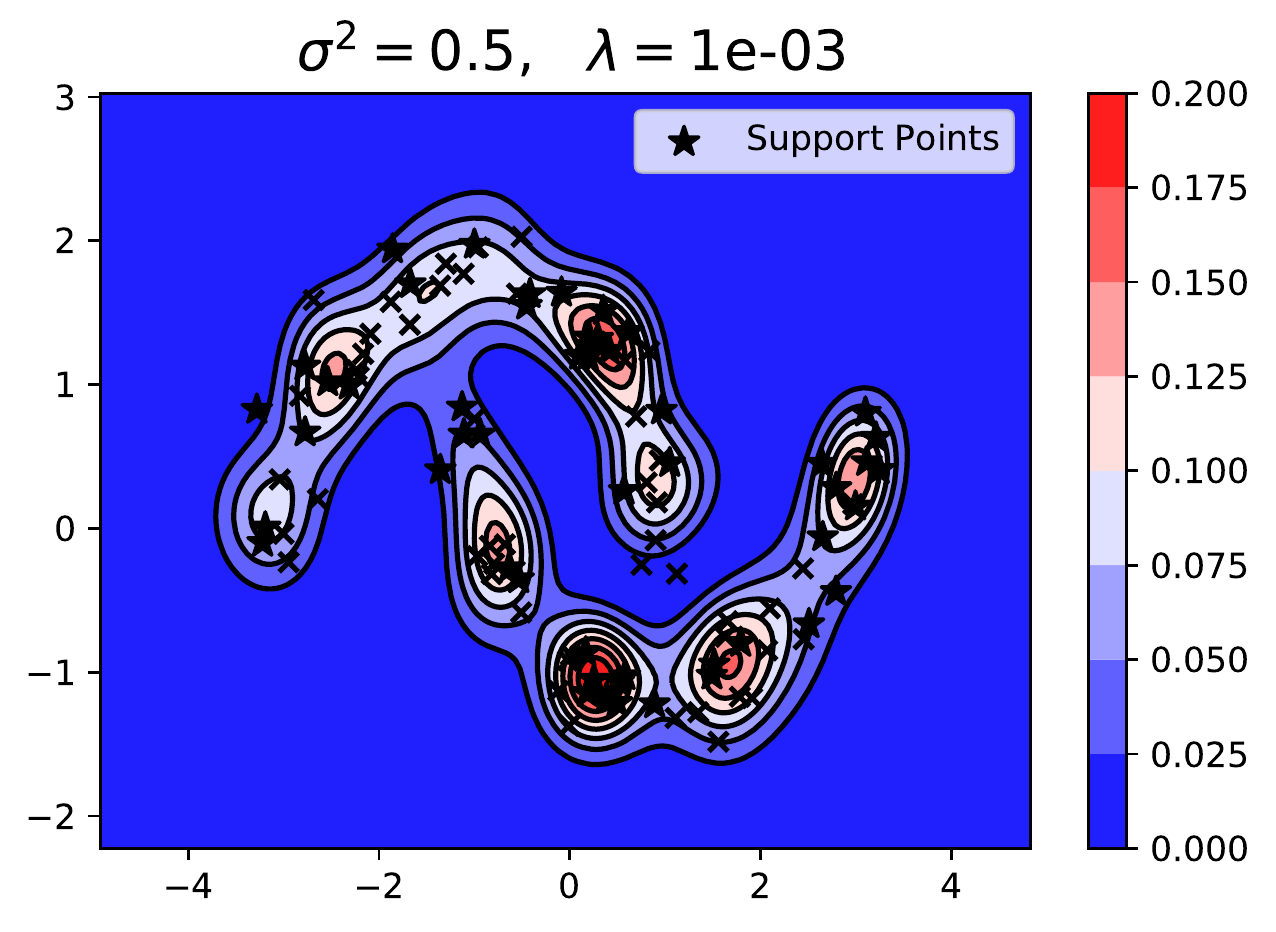}
    \includegraphics[width = .32\textwidth]{figs/density/moons_n_100_m_50_sigma_1.0_lambda_0.001_trace.pdf}
    \includegraphics[width =  .32\textwidth]{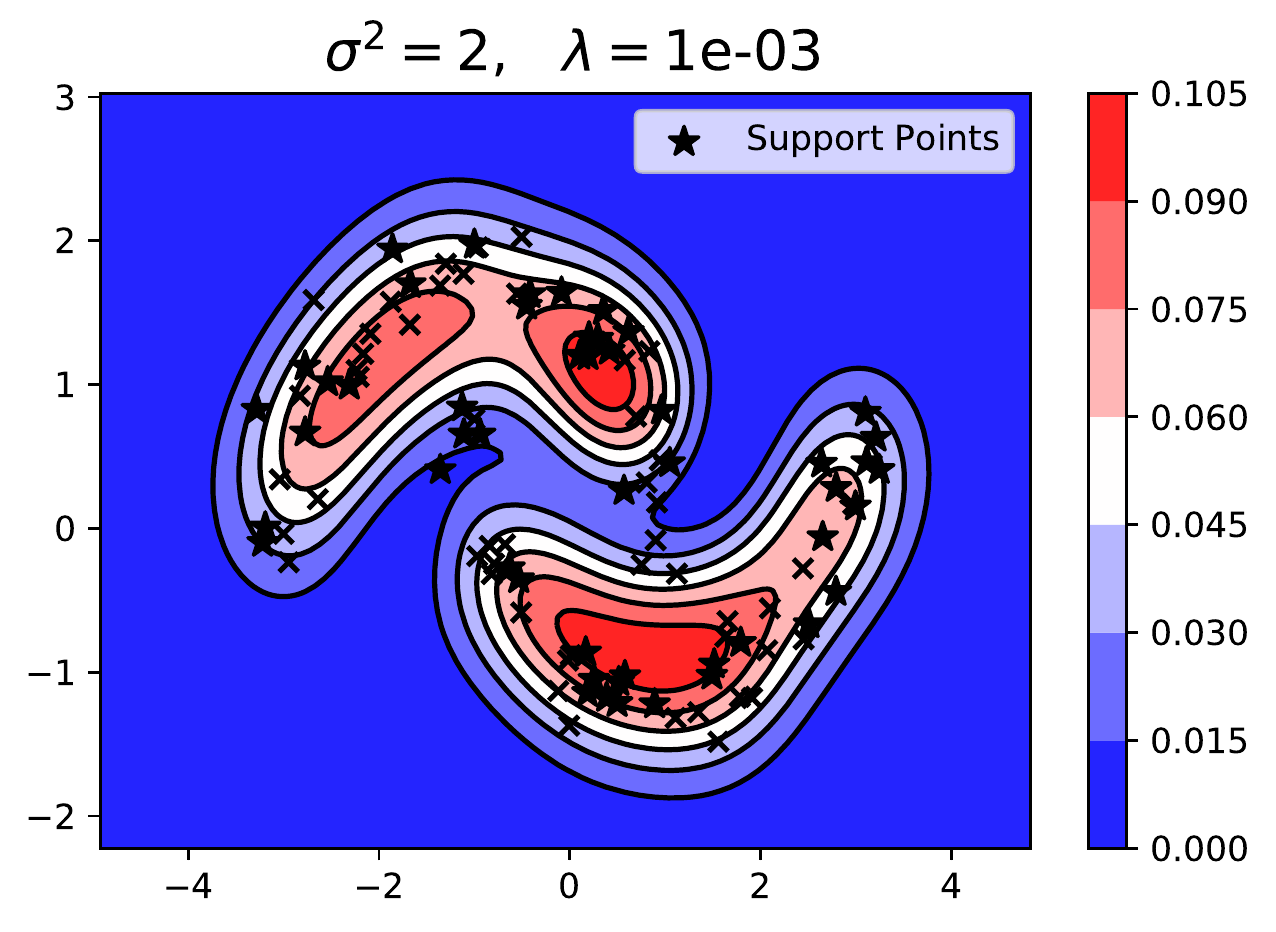}
    \caption{Effect of bandwidth.}
    \label{fig:sigma_effect}
\end{subfigure}
\begin{subfigure}[b]{1.\textwidth}
    \centering
     \begin{subfigure}[b]{0.32\textwidth}
         \centering
         \includegraphics[width =\textwidth]{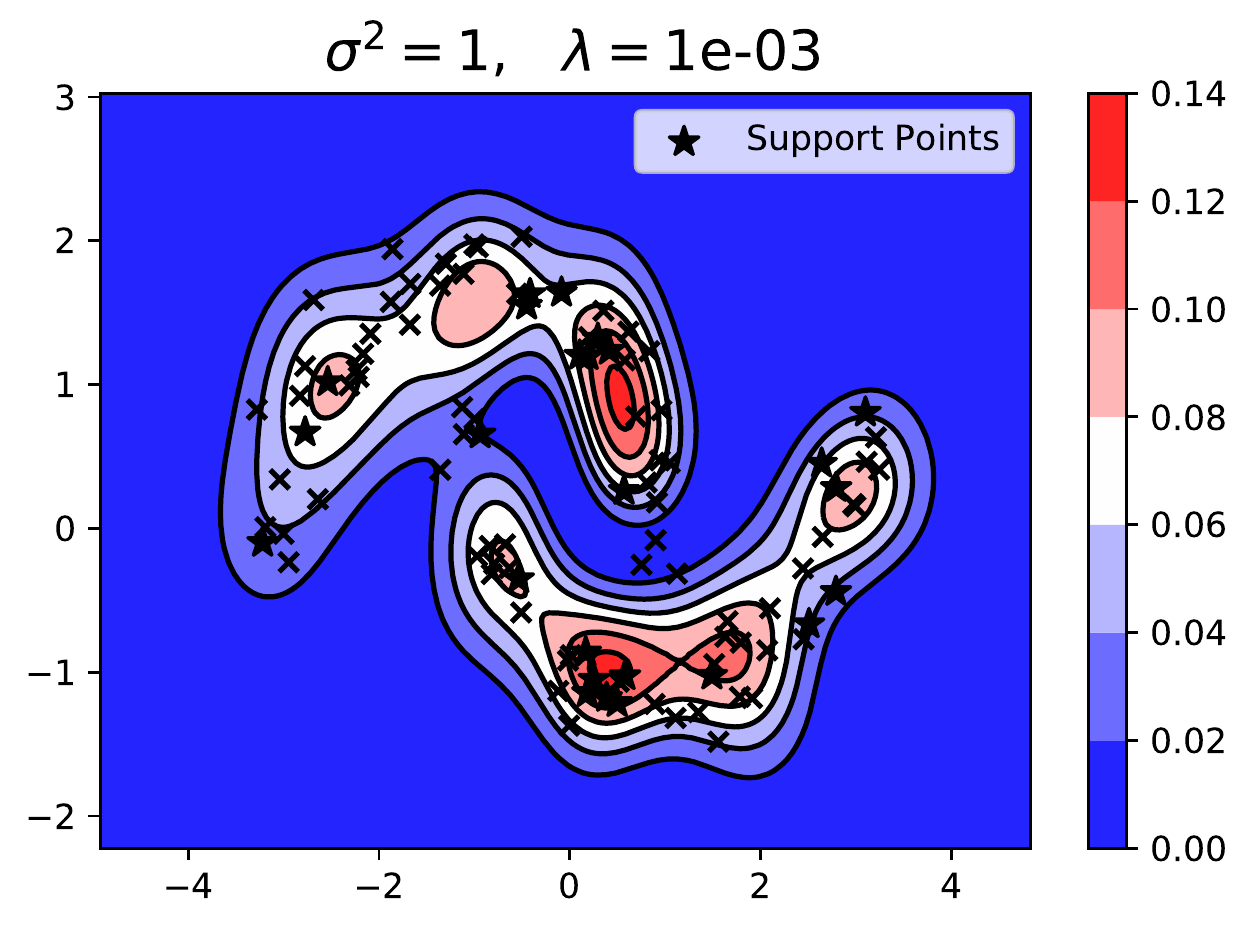}
         \caption*{$m=25$}
     \end{subfigure}
     \hfill
          \begin{subfigure}[b]{0.32\textwidth}
         \centering
         \includegraphics[width =\textwidth]{figs/density/moons_n_100_m_50_sigma_1.0_lambda_0.001_trace.pdf}
         \caption*{$m=50$}
     \end{subfigure}
     \hfill
          \begin{subfigure}[b]{0.32\textwidth}
         \centering
         \includegraphics[width =\textwidth]{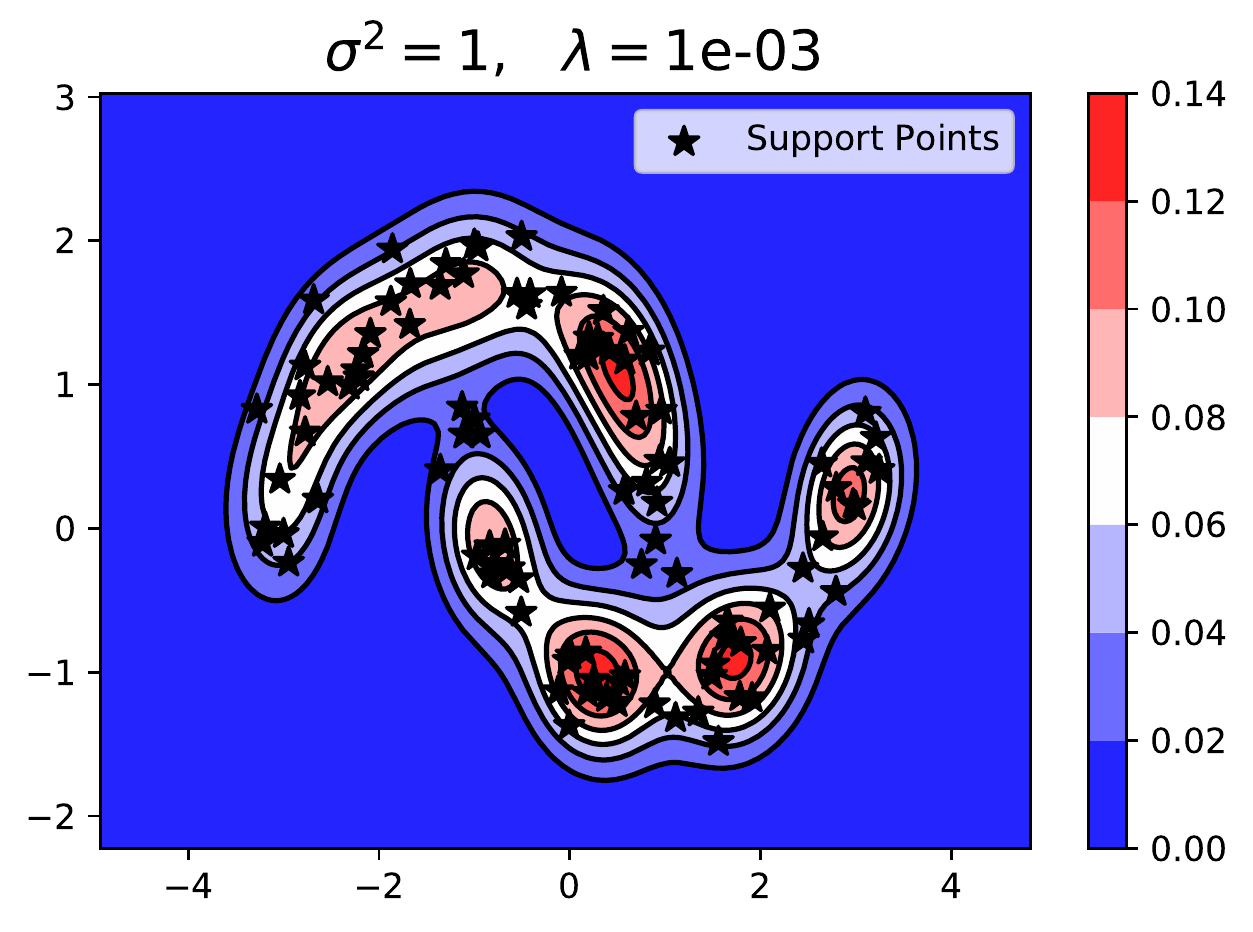}
         \caption*{$m=100$}
     \end{subfigure}
    \caption{Effect of sample size v.s. support size: $m$ out of $n=100$ supporting points are sampled.}
    \label{fig:size_effect}
\end{subfigure}
\end{figure}
We provide numerical results of a 2D density fitting task. We leave to future work the implementation of more complex applications, such as the DRO problem described above.
In \Cref{fig:lbda_effect,fig:sigma_effect,fig:size_effect}, we sample $n=100$ points from a ``two moons'' distribution, out of which $m$ points are used as support points $\tx_1, ..., \tx_m$, and minimize \cref{eq:densit_fitting} using the Gaussian kernel. In \Cref{fig:lbda_effect,fig:sigma_effect}, we fix $m=50$ and we illustrate the effect of bandwidth v.s.\ trace regularization, while in \Cref{fig:size_effect} we show the impact of varying the support size $m$. The code to reproduce these figures is available at \url{https://github.com/BorisMuzellec/kernel-SoS-distributions}.

\section*{Conclusion and future work}
In this note, we proposed to represent smooth probability distributions using kernel sums-of-squares as a way to address the intractability of optimizing distributions in the MMD geometry. We showed that this representation is dense for the weak topology, and can therefore be used to approximate arbitrarily well the solution of such problems on the whole space of probability measures. Finally, we provided efficient algorithms to fit kernel sum-of-squares densities. We leave to future work the application of this model to more complex tasks, such as distributionally robust optimization.

\section*{Acknowledgments}
This work was funded in part by the French government under management of Agence Nationale de la Recherche as part of the “Investissements d’avenir” program, reference ANR-19-P3IA-0001(PRAIRIE 3IA Institute). We also acknowledge support from the European Research Council (grants SEQUOIA 724063 and REAL 947908), and support by grants from Région Ile-de-France.

%\clearpage
%%%%%%%%%%% Biblio %%%%%%%%%%%%%%%%%

\bibliographystyle{plainnat}  
\bibliography{references}

%%%%%%%%%%%%%%%%%%%%%%%%%%%%%%%%%%%%%%%%%%%%%%%%%%%%%%%%%%%%

\appendix
\section{Appendix}\label{sec:proofs}

\subsection{Proof of \Cref{lemma:density}}

\begin{proof}
    By definition it holds $\forall x \in \Xcal, p_A(x) \geq 0$. Hence, $p_A$ is a density if and only if $\int_\Xcal p_A(x)\dd\rho(x) = 1$.
    We have 
    \begin{align*}
            \int_\Xcal p_A(x)\dd\rho(x) &= \int_\Xcal \dotp{\phi(x)}{A\phi(x)}_\Hcal \dd\rho(x)\\
            &= \dotp{A}{\int_\Xcal \phi(x)\otimes \phi(x) \dd\rho(x)}_{\HS}\\
            &= \dotp{A}{\Sigma_\rho}_\HS,
    \end{align*}
    where $\dotp{\cdot}{\cdot}_\HS$ denotes the Hilbert-Schmidt inner product, and the operator $\Sigma_\rho \defeq \int_\Xcal \phi(x)\otimes \phi(x) \dd\rho(x)$ is well-defined from the assumption $\sup_{x\in \Xcal} k(x, x) < \infty$\footnote{Note that the weaker assumption $\int_{x\in \Xcal} k(x, x)^2\dd\rho(x) < \infty$ would be enough.}.
\end{proof}

\subsection{Proof of \Cref{prop:weak_density}}

\begin{proof}
We consider w.l.o.g.\ the case where $\rho$ is the Lebesgue measure on $\Xcal$. Since $\mathcal{AC}(\Xcal)$ is dense in $\Prob(\Xcal)$ for the weak topology, it suffices to show that absolutely continuous probability measures with densities in $\Fcal_\rho$ are dense in $\mathcal{AC}(\Xcal)$.

Let $\mu \in \mathcal{AC}(\Xcal)$ with density function $f$. By compacity of $\Xcal$, using \Cref{thm:marteau_universality} we may construct a sequence $(A_n)_{n\in \NN} \in \pdm{\hh}^\NN$ with finite trace such that $p_{A_n}$ converges to $f$ almost-everywhere. Further, since $f$ is a density function, we may assume w.l.o.g.\ that $\forall n \in \NN, \tr A_n\Sigma_\rho=1$. By Scheffé's lemma \citep{scheffe1947useful}, we then have that $p_{A_n} \dd x$ converges in distribution to $\mu$, which implies weak convergence. This concludes the proof. 

\end{proof}

\subsection{Proof of \Cref{lemma:rkhs_norm}}

\begin{proof}
Let $A\in \pdm{\hh}, \tr A < \infty$ and $v \in \hh$. It holds 
\begin{align*}
    \|v - w_{p_A}\|^2_\hh = \|v\|_\hh^2 + \|w_{p_A}\|_\hh^2 - 2 \dotp{v}{w_{p_A}}_\hh.
\end{align*}
Since $w_{p_A} = \int \phi(x) \dotp{\phi(x)}{A\phi(x)}_\hh \dd\rho(x)$, we have
\begin{align*}
    \|w_{p_A}\|_\hh^2 &= \dotp{\int \phi(x) \dotp{\phi(x)}{A\phi(x)}_\hh\dd\rho(x)}{\int \phi(y) \dotp{\phi(y)}{A\phi(y)}_\hh\dd\rho(y)}_\hh\\
    &= \iint \dotp{\phi(x)}{\phi(y)}_\hh\dotp{\phi(x)}{A\phi(x)}_\hh\dotp{\phi(y)}{A\phi(y)}_\hh\dd\rho(x)\dd\rho(y)\\
    &= \iint k(x, y)\dotp{A}{\phi(x) \otimes \phi(x)}_\HS\dotp{A}{\phi(y) \otimes \phi(y)}_\HS\dd\rho(x)\dd\rho(y)\\
     &= \dotp{A}{\Tcal (A)}_\HS ~~\text{with}~~ \Tcal (A) \defeq \iint \phi(x)\otimes \phi(x) \dotp{\phi(y)}{A\phi(y)}_\hh k(x,y) \dd \rho(x) \dd\rho(y).
\end{align*}
Likewise, we have
\begin{align*}
    \dotp{v}{w_{p_A}}_\hh &= \dotp{v}{\int \phi(x) \dotp{\phi(x)}{A\phi(x)}_\hh\dd\rho(x)}_\hh\\
    &= \int\dotp{v}{\phi(x)}_\hh \dotp{\phi(x)}{A\phi(x)}_\hh\dd\rho(x)\\
    &= \int v(x) \dotp{A}{\phi(x)\otimes \phi(x)}_\HS\dd\rho(x)\\
    &= \dotp{A}{\int v(x)\phi(x)\otimes \phi(x)}_\HS\dd\rho(x)\\
     &= \dotp{A}{\Vcal}_\HS ~~\text{with}~~ \Vcal   \defeq \int v(x) \phi(x)\otimes \phi(x) \dd \rho(x).
\end{align*}
\end{proof}

\subsection{Proof of \Cref{lemma:mmd_closed_form}}

\begin{proof}
Let $\hh_m = \Span\{\phi(\tx_1), ..., \phi(\tx_m)\}$ and $A\in\pdm{\hh_m}$, there exists $\bB \in \pdm{\RR^m}$ such that $A = \sum_{i,j=1}^n B_{ij}\phi(\tx_i)\otimes\phi(\tx_j)$. From \Cref{lemma:density}, $p_A$ (and therefore $p_\bB$) is a density function if and only if $\tr A \Sigma_\rho = 1$. Since 
\begin{align*}
    \tr A\Sigma_\rho &= \sum_{i,j=1} B_{ij} \dotp{\phi(\tx_i)\otimes\phi(\tx_j)}{\int \phi(x) \otimes \phi(x) \dd \rho(x)}_\HS\\
    &= \sum_{i,j=1} B_{ij} \int \dotp{\phi(\tx_i)\otimes\phi(\tx_j)}{ \phi(x) \otimes \phi(x) }_\HS\dd \rho(x)\\
    &= \sum_{i,j=1} B_{ij} \int \dotp{\phi(\tx_i)}{\phi(x)}_\hh\dotp{\phi(\tx_j)}{\phi(x)}_\hh\dd \rho(x)\\
    &= \sum_{i,j=1} B_{ij} \int k(\tx_i, x)k(\tx_j, x)\dd \rho(x),
\end{align*}
this yields the equivalent condition $\tr \bB\bW = 1$ with $W_{ij} = \int k(\tx_i, x)k(\tx_j, x)\dd \rho(x), i,j \in [m].$

Let now $v = \sum_{i=1}^n a_i \phi(x_i)$, let us derive a finite-dimensional expression of $\|P_m(v - w_{p_A})\|^2_\Hcal,$
where $P_m$ is the projection operator on $\hh_m$. 
Let $\tilde{K}_{i,j} = k(\tx_i, \tx_j), i, j\in [m], K(\tilde{X}, X)_{ij} = k(\tx_i, x_j), i\in [m], j\in [n]$ and $\tilde{k}_x = [k(x, \tx_i)]_{i=1}^m$. $P_m$ satisifies $P_m(\phi(x)) = \sum_{i=1}^n \alpha_i \phi(\tx_i)$ with $\alpha = \tilde{\bK}\inv \tilde{k}_x$. For $p, q \in [m]$, let $\bu_{pq} = \int k(x, \tx_p) k(x, \tx_q) \tk_x \dd\rho(x) \in \RR^m$.
Since $A \in \pdm{\Hcal_m}$, $A$ can be written as $A = \sum_{ij} B_{ij}\phi(\tx_i)\otimes\phi(\tx_j)$ with $\bB \in \pdm{\RR^m}$. Hence, it holds
\begin{align*}
    P_m(v) &= \sum_{i=1} c_i \phi(\tx_i) ~~~\text{with} ~~~ c = \tilde{\bK}\inv \bK(\tilde{X}, X)a,
\end{align*}
and
\begin{align*}
    P_m(w_{p_A}) &= \sum_{pq} B_{pq} \int P_m(\phi(x))k(x, \tx_p) k(x, \tx_q) \dd\rho(x)\\
    &= \sum_{i,p,q }  B_{pq} \phi(x_i) \left(\tbK\inv \int  k(x, \tx_p) k(x, \tx_q)\tk_x \dd\rho(x)\right)_i\\
    &= \sum_{i=1} \beta_i \phi(\tx_i) ~~~\text{with} ~~~ \beta = \sum_{pq} B_{pq} \tilde{\bK}\inv \bu_{pq}.
\end{align*}
Further, we have
\begin{align*}
    \|P_m(v)\|_\Hcal^2 &=  a^T\bK(\tilde{X}, X)^T\tilde{\bK}\inv \bK(\tilde{X}, X)a,
\end{align*}
and
\begin{align*}
    \dotp{P_m(v)}{P_m(w_{p_A})}_\Hcal &= c^T\tilde{\bK} \beta\\
    &= \dotp{\bB}{\tilde{\bV}}_F ~~ \text{with} ~~ \tilde{V}_{pq} =  a^T \bK(\tilde{X}, X)^T \tilde{\bK}\inv \bu_{pq}\\
    &= c^T\Ucal(\bB),
\end{align*}
and finally
\begin{align*}
    \|P_m(w_{p_A})\|^2_\Hcal &= \beta^T\tilde{\bK} \beta\\
    &= \sum_{pqrs} B_{pq}B_{rs} \bu_{pq}^T \tilde{\bK}\inv \bu_{rs}\\
    &= \Ucal(\bB)^T\tilde{\bK}\inv\Ucal(\bB),
\end{align*}
where $\Ucal : \RR^{m\times m} \rightarrow \RR^m$ is the tensor of order 3 defined as $\Ucal(\bB) = \sum_{pq} B_{pq}\bu_{pq}$.

\end{proof}

\section{Closed forms in the Gaussian kernel}\label{sec:closed_forms}

In this section, we provide as an example closed forms for $\bW$ and $\bu_{pq}, p, q, \in [m]$ in the case where $\Xcal = \RR^d$, $\rho$ is the Lebesgue measure and $k(x, x') = e^{-\frac{\|x - x'\|^2}{\sigma^2}}$. Closed-forms for different kernels, supports and reference measures can be obtained in a similar way. 

\begin{lemma}
Let $\Xcal = \RR^d$, $\rho$ be the Lebesgue measure on $\RR^d$ and $\forall x, x' \in \RR^d, k(x, x') = e^{-\frac{\|x - x'\|^2}{\sigma^2}}$. For $p, q, r\in [m]$, we have
\begin{align*}
    W_{pq} &= \left(\frac{\pi\sigma^2}{2}\right)^{d/2}e^{-\frac{\|x_p - x_q\|^2}{2\sigma^2}}\\
    u_{pqr} &= \left(\frac{\pi\sigma^2}{3}\right)^{d/2} e^{-\frac{1}{\sigma^2}\left(\|x_p\|^2 + \|x_q\|^2 + \|x_r\|^2 - \tfrac{1}{3}\|x_p + x_q + x_r\|^2\right)}.
\end{align*}
\end{lemma}

\begin{proof}
Let $p, q \in [m]$. Since $\|x - x_p\|^2 + \|x - x_p\|^2 = 2 \|x - \frac{x_p + x_q}{2}\|^2 + \frac{1}{2}\|x_p - x_q\|^2$, we have 
\begin{align*}
    W_{pq} &\defeq \int_{\RR^d} k(x, x_p) k(x, x_q) \dd x\\
    &= \int_{\RR^d} e^{-\frac{\|x - x_p\|^2}{\sigma^2}}e^{-\frac{\|x - x_q\|^2}{\sigma^2}}\dd x\\
    &= e^{-\frac{\|x_p - x_q\|^2}{2\sigma^2}} \int_{\RR^d} e^{-\frac{2}{\sigma^2}\|x - \frac{x_p + x_q}{2}\|^2}\dd x\\
    &= \left(\frac{\pi\sigma^2}{2}\right)^{d/2} e^{-\frac{\|x_p - x_q\|^2}{2\sigma^2}}.
\end{align*}

Likewise, for $p, q, r \in [m]$ we have $$\|x - x_p\|^2 + \|x - x_p\|^2 + \|x - x_r\|^2 = 3 \left\|x - \frac{x_p + x_q + x_r}{3}\right\|^2 - \frac{1}{3}\|x_p + x_q + x_r\|^2 + \|x_p\|^2 + \|x_q\|^2 + \|x_r\|^2.$$ Hence, it holds 
\begin{align*}
    u_{pqr} &\defeq \int_{\RR^d} k(x, x_p) k(x, x_q) k(x, x_r) \dd x\\
    &= \int_{\RR^d} e^{-\frac{\|x - x_p\|^2}{\sigma^2}}e^{-\frac{\|x - x_q\|^2}{\sigma^2}}e^{-\frac{\|x - x_r\|^2}{\sigma^2}}\dd x\\
    &= e^{-\frac{1}{\sigma^2}\left(\|x_p\|^2 + \|x_q\|^2 + \|x_r\|^2 - \frac{1}{3}\|x_p + x_q + x_r\|^2  \right)} \int_{\RR^d} e^{-\frac{3}{\sigma^2}\left\|x - \frac{x_p + x_q + x_r}{3}\right\|^2}\dd x\\
    &= \left(\frac{\pi\sigma^2}{3}\right)^{d/2} e^{-\frac{1}{\sigma^2}\left(\|x_p\|^2 + \|x_q\|^2 + \|x_r\|^2 - \tfrac{1}{3}\|x_p + x_q + x_r\|^2\right)}.
\end{align*}

\end{proof}

\end{document}